\documentclass[conference]{IEEEtran}
\usepackage{cite}
\usepackage{amsmath,amssymb,amsfonts}
\usepackage{algorithmic}
\usepackage{graphicx}
\usepackage{textcomp}
\usepackage{booktabs} 
\usepackage{xcolor}
\usepackage{multirow} 
\def\BibTeX{{\rm B\kern-.05em{\sc i\kern-.025em b}\kern-.08em
    T\kern-.1667em\lower.7ex\hbox{E}\kern-.125emX}}
\begin{document}

\title{
Improving k-Means Clustering Performance with Disentangled Internal Representations
}

\author{\IEEEauthorblockN{Abien Fred Agarap}
\IEEEauthorblockA{\textit{College of Computer Studies} \\
\textit{De La Salle University}\\
Manila, Philippines \\
abien\_agarap@dlsu.edu.ph}
\and
\IEEEauthorblockN{Arnulfo P. Azcarraga}
\IEEEauthorblockA{\textit{College of Computer Studies} \\
\textit{De La Salle University}\\
Manila, Philippines \\
arnulfo.azcarraga@dlsu.edu.ph}}

\maketitle

\begin{abstract}
Deep clustering algorithms combine representation learning and clustering by jointly optimizing a clustering loss and a non-clustering loss. In such methods, a deep neural network is used for representation learning together with a clustering network. Instead of following this framework to improve clustering performance, we propose a simpler approach of optimizing the \textit{entanglement} of the learned latent code representation of an autoencoder. We define \textit{entanglement} as how close pairs of points from the same class or structure are, relative to pairs of points from different classes or structures. To measure the entanglement of data points, we use the \textit{soft nearest neighbor loss}, and expand it by introducing an annealing temperature factor. Using our proposed approach, the test clustering accuracy was 96.2\% on the MNIST dataset, 85.6\% on the Fashion-MNIST dataset, and 79.2\% on the EMNIST Balanced dataset, outperforming our baseline models.
\end{abstract}


\section{Introduction and Related Works}
Clustering is an unsupervised learning task that groups a set of objects in a way that the objects in a group share more similarities among them than those from other groups. It is a widely-studied task as its applications include but are not limited to its use in data analysis and visualization, anomaly detection, sequence analysis, and natural language processing. Like other machine learning methods, clustering algorithms heavily rely on the choice of feature representation. For this reason, the design of preprocessing pipelines and feature transformations (a.k.a. \textit{feature engineering}) consumes a considerable amount of time. In turn, the task of feature engineering plays a crucial role in the success of machine learning methods. However, due to its labor-intensive nature, it hinders faster development, autonomous learning, and reusability across tasks.

We take $k$-means clustering as an example, which typically uses the Euclidean distance among points in a given feature space (e.g. for images, it could be the raw pixels or gradient-orientation histograms); for difficult image datasets like CIFAR10\cite{krizhevsky2009learning}, clustering with Euclidean distance on raw pixels may be ineffective.

Hence, the move towards automatically learning the best representation for a given data has gained mainstream attention since the success of deep learning for computer vision\cite{he2016deep, krizhevsky2012imagenet, simonyan2014very}, where the exceptional gains on benchmark tasks have resulted from automatically learning better feature representation. The task of automatically learning feature representation is known as \textit{representation learning}.

More explicitly, representation learning is the task of learning the most salient features of a given data, i.e. features that imply the underlying structure of the data. It is implicitly done in a supervised deep neural network by using its hidden layers to learn and to provide representations for its last layer, thereby rendering a task such as classification or regression easier. For instance, data points that are not linearly separable in the raw feature space may become linearly separable at the last hidden layer through the composition of feature representations in the hidden layers. So, by automatically learning the representations instead of feature engineering, we are unencumbered of the task to learn the best possible feature representation for better performance in downstream tasks such as classification, clustering, and regression.
To further take advantage of representation learning, it may be explicitly designed to forge representations in favor of a downstream task such as clustering. In the following subsections, we briefly discuss related works that use the aforementioned strategy.

\subsection{Deep Embedded Clustering (DEC)}
Xie et al. (2016)\cite{xie2016unsupervised} introduced the Deep Embedded Clustering (DEC), a method that simultaneously learns feature representations and cluster assignments using a deep neural network. DEC learns a mapping $f_{\theta}: X \rightarrow Z$ where $X$ is the original feature space, while $Z$ is the lower-dimensional feature space. Then, it iteratively optimizes the Kullback-Leibler divergence to minimize the within-cluster distance of each cluster in $Z$. They had a clustering accuracy of 84.30\% on the MNIST dataset\cite{lecun1998gradient}.

\subsection{Variational Deep Embedding (VaDE)}
Another approach to strongly influence learned representations to favor clustering is the Variational Deep Embedding (VaDE) by Jiang et al. (2016)\cite{jiang2016variational}. It is an unsupervised generative clustering approach that employs the framework of Variational Autoencoder\cite{kingma2013autoencoding}, by combining a Gaussian Mixture Model (GMM) and a deep neural network (DNN). Specifically, a cluster is chosen by the GMM from which the latent representation $z$ is sampled, and then the DNN decodes $z$ to an observable data $x$. Their approach had a clustering accuracy of 94.46\% on the MNIST dataset\cite{lecun1998gradient}.

\subsection{ClusterGAN}
Similar to DEC\cite{xie2016unsupervised} and VaDE\cite{jiang2016variational}, ClusterGAN\cite{mukherjee2019clustergan} also uses learned representation for clustering, but their approach built on the Generative Adversarial Network (GAN)\cite{goodfellow2014generative} framework. That is, they incorporated a clustering-specific loss term to the minimax objective. They had a clustering accuracy of 95\% on the MNIST dataset\cite{lecun1998gradient} and 63\% on the Fashion-MNIST dataset\cite{xiao2017fashion}.

\subsection{N2D: (not too) deep clustering}
DEC\cite{xie2016unsupervised}, VaDE\cite{jiang2016variational}, and ClusterGAN\cite{mukherjee2019clustergan} are all deep clustering algorithms, i.e. methods that combine a deep neural network as a data encoder $f_{\theta}: X \rightarrow Z$, and a clustering network, that jointly learns a clustering loss and a non-clustering loss (e.g. minimax objective for GAN). In contrast, McConville et al. (2019)\cite{mcconville2019n2d} proposed N2D (Not too Deep) clustering, wherein they performed manifold learning on the latent code representation from an autoencoder, and used the learned manifold for clustering. They found that using UMAP in their framework achieves the best clustering-friendly manifold of the latent code representation. Using this approach, they had a clustering accuracy of 94.8\% on the MNIST dataset\cite{lecun1998gradient} and 67.2\% on the Fashion-MNIST dataset\cite{xiao2017fashion}.

Similar to N2D\cite{mcconville2019n2d}, we propose a relatively simpler approach compared to deep clustering algorithms. But we also use an autoencoder to learn the latent code representation of a data, then use the said representation for clustering. We draw our difference on how to learn a more clustering-friendly latent code representation.

Instead of learning such a representation by using a manifold learning technique such as Isomap\cite{tenenbaum2000global}, t-SNE\cite{maaten2008visualizing}, or UMAP\cite{mcinnes2018umap}, we regularize the autoencoder reconstruction loss with the \textit{soft nearest neighbor loss}\cite{salakhutdinov2007learning, frosst2019analyzing}. The said loss function measures the lack of separation of class (or structural, for unsupervised tasks) manifolds in representation space--in other words, the \textit{entanglement} of data points from different classes or structures.

We focus on minimizing the entanglement of class manifolds in a latent code representation to derive a more clustering-friendly representation.

Our contributions are as follows:
\begin{enumerate}
    \item We expand the soft nearest neighbor loss by introducing an annealing temperature factor, and we use it to learn a latent code representation that is primed for clustering (Section \ref{section:learning_disentangled_representations}).
    \item We present comparatively strong clustering results in terms of clustering accuracy, normalized mutual information, and adjusted Rand index (Section \ref{section:clustering_disentangled_representations}).
    \item A simple yet effective way of clustering on (disentangled) latent code representation (Section \ref{section:clustering_disentangled_representations}).
\end{enumerate}

\section{Learning Disentangled Representations}\label{section:learning_disentangled_representations}
We consider the problem of clustering a set of $N$ points $\{x_{i} \in X\}_{i=1}^{N}$ into $k$ clusters, each represented by a centroid $\mu_{j\in1,\ldots,k}$. Instead of directly clustering the original features $X$, we transform the data with a non-linear mapping $Z = enc(X)$, where $Z$ is the latent code representation. But to learn a more clustering-friendly representation, we propose to learn to \textit{disentangle} them, i.e. isolate class- or structure-similar data points, which implicitly maximizes the inter-cluster variance.

\subsection{Autoencoder}
An autoencoder is a neural network that aims to find the function mapping the features $x$ to itself through the use of an encoder function $h = enc(x)$ that learns the latent code representation of the features, and a decoder function that reconstructs the features from the latent code representation $r = dec(h)$. To learn the reconstruction task, it minimizes a loss function $\mathcal{L}(x, dec(enc(x)))$, where $\mathcal{L}$ is a function penalizing the decoder output $dec(enc(x))$ for being dissimilar from $x$. Typically, this reconstruction loss is the Mean Squared Error (MSE) $\frac{1}{n} \sum_{i=1}^{n}\|dec(enc(x_{i})) - x_{i}\|_{2}^{2}$. Then, similar to other neural networks, it is usually trained with a gradient-based method aided with backpropagation of errors.

The reconstruction task of an autoencoder has a by-product of learning good internal representations, and so, we take advantage of this for clustering, particularly, the latent code representation $z = enc(x)$ -- thus drawing our similarity with N2D\cite{mcconville2019n2d}. For our experiments, we used the binary cross entropy (Eq. \ref{eq:sigmoid_recon_loss}) as the reconstruction loss.
\begin{equation}\label{eq:sigmoid_recon_loss}
    \ell_{rec}(x, r) = \dfrac{1}{n} \sum_{i=1}^{n} -x_{i}\log(r_{i}) + (1 - x_{i})\log(1 - r_{i})
\end{equation}
We used this in lieu of MSE since our features $X$ were normalized to values $[0, 1]\in\mathbb{R}$.
\subsection{Soft Nearest Neighbors Loss}
In our context, we define \textit{entanglement} as how close pairs of representations from the same class or structure are, relative to pairs of representations from different classes or structures. Frosst et al. (2019)\cite{frosst2019analyzing} used the same term in the same context. A low entanglement implies that representations from the same class or structure are closer than they are to representations from different classes or structures. To measure the entanglement of representations, Frosst et al. (2019)\cite{frosst2019analyzing} expanded the non-linear neighborhood component analysis (NCA)\cite{salakhutdinov2007learning} objective by introducing the temperature factor $T$, and called this modified objective the \textit{soft nearest neighbor loss}.

They defined the \textit{soft nearest neighbor loss} as the non-linear NCA at temperature $T$, for a batch of $b$ samples $(x, y)$,
\begin{equation}\label{eq:snnl}
    \ell_{sn}(x, y, T) = -\dfrac{1}{b} \sum_{i \in 1 \dots b} \log \left( \dfrac{\sum\limits_{\substack{j \in 1 \dots b \\ j \neq i \\ y_{i} = y_{j}}} e^{-\frac{\|x_{i} - x_{j}\|^{2}}{T}} }{\sum\limits_{\substack{k \in 1 \dots b \\ k \neq i}} e^{-\frac{\|x_{i} - x_{k}\|^{2}}{T}}} \right)
\end{equation}

where $x$ may be the raw input features or the learned representations in the hidden layers of a neural network. We may describe soft nearest neighbor loss as the negative log probability of sampling a neighboring point \textit{j} from the same class as \textit{i} in a batch \textit{b}, similar to the probabilistic sampling by Goldberger et al. (2005)\cite{goldberger2005neighbourhood}. A low value of soft nearest neighbor loss is tantamount to a low entanglement (high disentanglement). In our experiments, we used cosine similarity instead of Euclidean distance.
\subsubsection{Temperature} Frosst et al. (2019)\cite{frosst2019analyzing} described the temperature factor \textit{T} as a way to control the relative importance given to the distances between pairs of points, i.e. at low temperatures, the loss is dominated by small distances while the actual distances between widely separated representations become less relevant. Conversely, at high temperatures, the distances between widely separated points dominate the loss.

\subsubsection{Annealing Temperature} 
\begin{figure}[htb!]
    \centering
    \includegraphics[width=\linewidth]{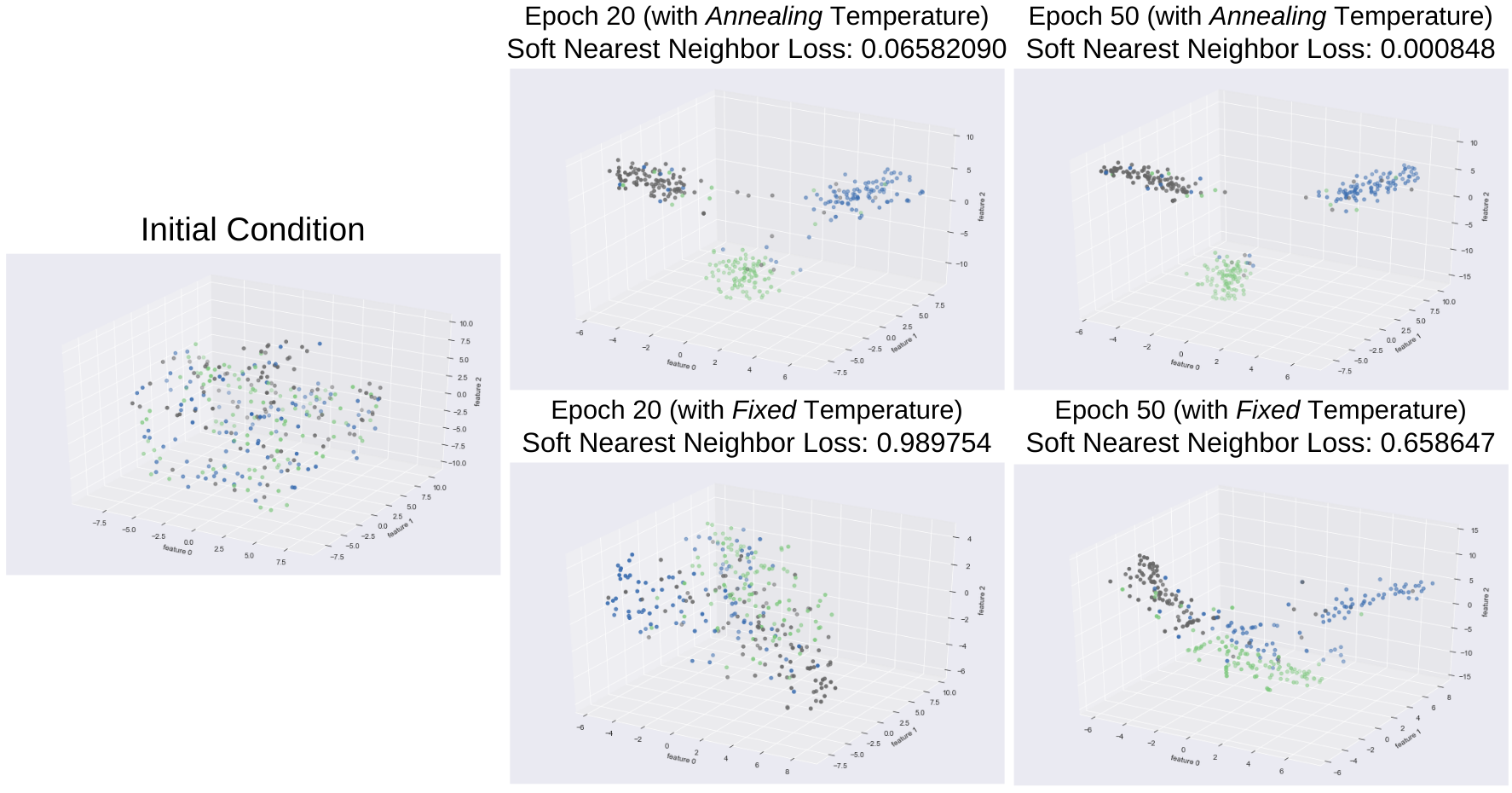}
    \caption{Comparing the soft nearest neighbor loss with annealing temperature and with fixed temperature. We sampled and randomly labelled 300 data points from a Gaussian distribution, and ran gradient descent on them with soft nearest neighbor loss. The figure at the left shows the initial condition of the labelled points. We can see the separation of clusters in the latent code from epoch 20 to epoch 50, rendering the classes more isolated. We present disentangled representations on benchmark datasets later in the paper. This figure is best viewed in color.}
    \label{fig:synthetic_snnl}
\end{figure}

We build on this idea of the temperature influencing the importance of distances between pairs of points, and extend the soft nearest neighbor loss by introducing an annealing temperature (Eq. \ref{eq:annealing_temp}) instead of a fixed one,
\begin{equation}\label{eq:annealing_temp}
    T = \dfrac{1}{(\eta + i)^{\gamma}}
\end{equation}
where $i$ is the current training epoch, and we set $\eta=1$ and $\gamma=0.55$ for our experiments, similar to Neelakantan et al. (2015)\cite{neelakantan2015adding}. We show a disentangled latent code representation learned with the soft nearest neighbor loss, both with fixed temperature and with annealing temperature, in Figure \ref{fig:synthetic_snnl}. In both cases, as we minimize the soft nearest neighbor loss, the latent code representation becomes more clustering-friendly. However, with annealing temperature, we gain a lower entanglement at an earlier training epoch, thus the disentanglement of representations starts earlier than with a fixed temperature.

We also used the lowest soft nearest neighbor loss among the hidden layers of our autoencoder:
\begin{equation}
    \ell'_{sn} = \arg\min \ell_{sn}(x, y, T)
\end{equation}
Using the $\arg\min$ configuration, we achieved a more stable computation of the soft nearest neighbor loss during training.

Now that we have defined our contribution, we lay down the objective function used by our autoencoder to learn the disentangled representations for clustering. We define a composite loss (Eq. \ref{eq:composite_loss}) that consists of the autoencoder reconstruction loss $\ell_{rec}$, and the soft nearest neighbor loss $\ell_{sn}$, with an $\alpha$ parameter which directly influences the disentanglement computation -- we set $\alpha=100$. Note that our goal is not necessarily to learn a good reconstruction of the input data, but to learn a good disentangled representation that we can use to improve clustering performance.
\begin{equation}\label{eq:composite_loss}
    \mathcal{L}(f, x, y) = \ell_{rec}(x, r) + \alpha \cdot \sum_{i \in k} \ell_{sn}\left(f^{i}(x), y\right)
\end{equation}

Another contribution of this work is the simplicity of our proposed method. So, unlike DEC\cite{xie2016unsupervised}, VaDE\cite{jiang2016variational}, and ClusterGAN\cite{mukherjee2019clustergan} which uses an auxiliary clustering network, we use a simple k-Means clustering\cite{lloyd1982least} on the disentangled latent code representations.

Thus far, we have discussed our proposed method of disentangling learned representations to improve clustering performance. However, autoencoding and clustering are both unsupervised learning tasks, while we are proposing to use the soft nearest neighbor loss, a loss function that uses labels to illuminate the class similarity structure of internal representations learned by a neural network. With this in mind, we formulated two different soft nearest neighbor loss functions: (1) supervised, and (2) unsupervised. In the unsupervised setting, we simply perform the same probabilistic sampling of a neighboring point \textit{j} to \textit{i}, but we do not constrain them to come from the same class.

To simulate the lack of labelled data, we used a small labelled subset of the benchmark datasets for the supervised configuration of the soft nearest neighbor loss. The different soft nearest neighbor loss configurations we used in our experiments are listed in Table \ref{tab:model_lookup}.

\begin{table}[htb]
    \caption{Configurations of Autoencoder (AE) Trained with Soft Nearest Neighbor Loss (SNNL)}
    \label{tab:model_lookup}
    \centering
    \begin{tabular}{c|l}
        \toprule
         Shorthand & Full SNNL-trained Autoencoder Configuration \\ 
        \midrule
         SNNL-1 & Supervised SNNL-trained AE w/ fixed $T$ \\
         SNNL-2 & Unsupervised SNNL-trained AE w/ fixed $T$ \\
         SNNL-3 & Supervised SNNL-trained AE w/ $\arg\min$ and fixed $T$\\
         SNNL-4 & Unsupervised SNNL-trained AE w/ $\arg\min$ and fixed $T$\\
         SNNL-5 & Supervised SNNL-trained AE w/ annealing $T$\\
         SNNL-6 & Unsupervised SNNL-trained AE w/ annealing $T$\\
         SNNL-7 & Supervised SNNL-trained AE w/ $\arg\min$ and annealing $T$\\
         SNNL-8 & Unsupervised SNNL-trained AE w/ $\arg\min$ and annealing $T$\\
        \bottomrule
    \end{tabular}
\end{table}

We use Table \ref{tab:model_lookup} as a lookup table of shorthand to save space on Tables \ref{tab:mnist_clustering_performance_latent_code}, \ref{tab:fmnist_clustering_performance_latent_code}, and \ref{tab:emnist_clustering_performance_latent_code}. Model configurations labelled as SNNL 1--4 use a fixed temperature, while those that are labelled as SNNL 5--8 use our annealing temperature (Eq. \ref{eq:annealing_temp}). Moreover, model configurations with even numbers use the unsupervised soft nearest neighbor loss, while the ones with odd numbers use the supervised version.

Finally, we summarize the details of our proposed method as follows,

\begin{enumerate}
    \item Train an autoencoder with the composite loss (Eq. \ref{eq:composite_loss}) using a gradient-based method with backpropagation.
    \item Use the disentangled latent code representation $z = enc(x)$ of the autoencoder for k-Means clustering.
\end{enumerate}

\section{Clustering on Disentangled Representations}\label{section:clustering_disentangled_representations}
To demonstrate the effectiveness of our approach, we conducted experiments on benchmark datasets, and lay down the clustering performance of the related models on the same benchmark datasets we used.

\subsection{Autoencoder Model}
We used a fully connected network with $d-500-500-2000-c-2000-500-500-\hat{d}$ units, where $d$ and $\hat{d}$ (s.t. $d = \hat{d}$) refer to the dimensionality of the features, and $c$ refers to the dimensionality of the latent code representation -- similar to the one used by Salakhutdinov and Hinton (2007)\cite{salakhutdinov2007learning}. The $c$-latent code layer and $\hat{d}$-reconstruction layer used logistic function while the remaining hidden layers used ReLU function\cite{nair2010rectified}. All hidden layers were initialized with He initialization\cite{he2015delving}. We set $c = 70$ across all datasets and models for a more fair comparison. Finally, we trained our model using Adam\cite{kingma2014adam} with a learning rate of $1\times10^{-3}$ for 50 epochs on all our datasets.

\subsection{k-Means Clustering}
We ran k-Means clustering, with centroids initialized using \texttt{k-means++}\cite{arthur2007kmeans}, on the disentangled latent code representation from our autoencoder for nine times. The first run started with 10 iterations, with each run incremented by 10, i.e. the first clustering ran for 10 iterations, while the ninth ran for 90 iterations. We recorded the clustering performance on the ninth run for each model on each dataset.

\subsection{Evaluation Metrics}
For all the models, we used the number of ground-truth categories in each dataset as the number of clusters. We used six different metrics to evaluate the clustering performance of our baseline and experimental models. For the first three metrics, the values lie in the interval $[0, 1]$, where values closer to 1 correspond to a better clustering performance. The values for the fourth metric lie in the interval $[-1, 1]$, where values near 1 are better while values near -1 are worse. The last two metrics are unbounded, but only the first of which implies a better clustering performance when its values are higher, while the last metric requires a lower value for better performance.

\subsubsection{Clustering Accuracy}
In clustering, accuracy (ACC) is defined as the best match between the ground-truth labels and the predicted clusters\cite{yang2010image}. Using the ground-truth labels as pseudo-cluster labels is known as \textit{cluster assumption} in the semi-supervised learning literature\cite{chapelle2009semisupervised}.
\begin{equation}
    ACC = \max_{m} \dfrac{\sum\limits_{i=1}^{n}1\{l_{i}=m\left(c_{i}\right)\}}{n},
\end{equation}

where $l_{i}$ is the ground-truth label, $c_{i}$ is the cluster prediction, and $m$ ranges over all possible one-to-one mappings between clusters and labels.

\subsubsection{Normalized Mutual Information}
The Normalized Mutual Information (NMI) is the normalization of mutual information (MI) score to have its value within $[0, 1]\in\mathbb{R}$, where 0 denotes no mutual information while 1 denotes perfect correlation. It is formally defined as follows,
\begin{equation}
    NMI = \dfrac{2I(y, c)}{[H(y) + H(c)]}
\end{equation}
where $y$ is the ground-truth label, $c$ is the cluster prediction, $H$ is the entropy, and $I$ is the mutual information between the ground-truth labels and the cluster predictions.

\subsubsection{Adjusted Rand Index}
The Adjusted Rand Index (ARI)\cite{hubert1985comparing} is the Rand Index (RI) adjusted for chance. RI is the similarity between two clusterings by considering all pairs of points and counting pairs assigned to the same or different clusters in the predicted and true clusterings (see Eq. \ref{eq:ri}). 
\begin{equation}\label{eq:ri}
    RI = \dfrac{TP + TN}{TP + FP + FN + TN}
\end{equation}
where $TP$ is the true positive, $TN$ is the true negative, $FP$ is the false positive, and $FN$ is the false negative. ARI is then computed from RI by using Eq. \ref{eq:ari},
\begin{equation}\label{eq:ari}
    ARI = \dfrac{RI - \mathbb{E}[RI]}{\max(RI) - \mathbb{E}[RI]}
\end{equation}
ARI values lie within $[0, 1]\in\mathbb{R}$, where 0 denotes random labelling independently of the number of clusters, while 1 denotes the clusterings are identical up to a permutation.

\subsubsection{Silhouette Score} The Silhouette Score (SIL)\cite{rousseeuw1987silhouettes} measures the similarity of examples to their own cluster compared to other clusters, and it is computed by using Eq. \ref{silhouette-score},
\begin{align}\label{silhouette-score}
    SIL = \dfrac{b(i) - a(i)}{\max\left[a(i), b(i)\right]}
\end{align}
where $a(i)$ is the distance between point $i$ and all other points in its cluster, and can be computed by using Eq. \ref{silhouette-score-a},
\begin{align}\label{silhouette-score-a}
    a(i) = \dfrac{1}{|C_{i}| - 1} \sum_{j \in C_{i}, i \neq j} d(i, j)
\end{align}
where $C_{i}$ is the predicted cluster for point $i$, and $d(i, j)$ is the distance between points $i$ and $j$.

Then, $b(i)$ is the distance between point $i$ and all other points in the next nearest cluster, and can be computed by using Eq. \ref{silhouette-score-b},
\begin{align}\label{silhouette-score-b}
    b(i) = \min_{k \neq i} \dfrac{1}{|C_{k}|} \sum_{j \in C_{k}} d(i, j)
\end{align}
Any distance metric may be used, but we used the Euclidean distance metric in our evaluation.

\subsubsection{Calinski-Harabasz Score} The Calinski-Harabasz score\cite{calinski1974dendrite} (CHS) is defined as the ratio between within-cluster dispersion and between-cluster dispersion, and it is computed by using Eq. \ref{ch-score}. A higher CHS implies better cluster separation.
\begin{align}\label{ch-score}
    CHS = \dfrac{Tr(B_{k})}{Tr(W_{k})} \times \dfrac{N - k}{k - 1}
\end{align}
where $B_{k}$ is the between-cluster dispersion matrix given by Eq. \ref{bk},
\begin{align}\label{bk}
    B_{k} = \sum_{q} n_{q} (c_{q} - c)(c_{q} - c)^{T}
\end{align}
and $W_{k}$ is the within-cluster dispersion given by Eq. \ref{wk},
\begin{align}\label{wk}
    W_{k} = \sum_{q = 1}^{k}\sum_{x \in C_{q}} (x - c_{q})(x - c_{q})^{T}
\end{align}
where $N$ is the number of points in a data, $C_{q}$ is the set of points in cluster $q$, $c_{q}$ is the center of cluster $q$, $c$ is the center of $C$, and $n_{q}$ is the number of points in cluster $q$.

\subsubsection{Davies-Bouldin Index} The Davies-Bouldin Index (DBI) is defined as the ratio of within-cluster distances to between-cluster distances\cite{davies1979cluster, halkidi2001clustering}. A lower DBI implies better cluster separation. It is computed by using Eq. \ref{db-index},
\begin{align}\label{db-index}
    DBI = \dfrac{1}{k} \sum_{i=1}^{k} \max_{i \neq j} R_{j}
\end{align}
\indent where $R$ is the average similarity among clusters given by Eq. \ref{average-similarity}.
\begin{align}\label{average-similarity}
    R_{ij} = \dfrac{s_{i} + s_{j}}{d_{ij}}
\end{align}
\indent where $s_{i}$ is the cluster diameter which is the average distance between each point in cluster $i$ and the cluster centroid, and $d_{ij}$ is the distance between centroids $i$ and $j$.

\subsection{Datasets Description}
We evaluate and compare our baseline and experimental methods on three image datasets. We list the dataset statistics in Table \ref{tab:data_stats}.
\begin{table}[htb]
    \caption{Datasets statistics.}
    \label{tab:data_stats}
    \centering
    \begin{tabular}{c|c|c|c}
        \hline
         Dataset & \# Samples & Input Dimension & \# Clusters\\
        \hline
         MNIST &  70,000 & 784 & 10\\
         Fashion-MNIST & 70,000 & 784 & 10\\
         EMNIST Balanced & 131,600 & 784 & 47\\
        \hline
    \end{tabular}
\end{table}
\begin{enumerate}
    \item MNIST: The MNIST handwritten digit classification dataset\cite{lecun1998gradient} consists of 60,000 training examples and 10,000 test examples -- all in grayscale, and of size 28 by 28 pixels. We reshaped each image to a 784-dimensional vector.
    \item Fashion-MNIST: The Fashion-MNIST\cite{xiao2017fashion} is a more challenging alternative to the MNIST dataset, which consists of 60,000 training examples and 10,000 test examples -- also all in grayscale, and of size 28 by 28 pixels. We reshaped each image to a 784-dimensional vector.
    \item EMNIST Balanced: The EMNIST Balanced is a subset of the EMNIST handwritten characters classification dataset\cite{cohen2017emnist}, that has a 47 balanced classes of handwritten digits and letters. It has 112,800 training examples and 18,800 test examples -- also all in grayscale, and of size 28 by  28 pixels. We reshaped each image to a 784-dimensional vector.
\end{enumerate}

For the supervised configuration of the soft nearest neighbor loss, we simulate the lack of labelled data by randomly picking 10,000 labelled training examples each for the MNIST and the Fashion-MNIST datasets, and randomly picking 20,000 labelled training examples for the EMNIST balanced dataset, since it has to be clustered in 47 groups. Despite using a small subset of labelled training examples, we still used the full test sets for the clustering task.

On the other hand, we used the full training sets for k-Means clustering on the original feature representation, and for the unsupervised learning models (i.e. baseline autoencoder, SNNL-\{2, 4, 6, 8\}), and we evaluated them on the full test sets.

\subsection{Clustering Performance}\label{section:clustering_performance}
We evaluate the performance of our experimental method in its different configurations. For our baseline models, we used k-Means clustering on (1) the original feature representation, encoded using principal components analysis (PCA) to 70 dimensions in order to avoid the ``curse of dimensionality''\cite{bellman1961adaptive}, and (2) on the latent code representation from an autoencoder trained with reconstruction loss (Eq. \ref{eq:sigmoid_recon_loss}) only. Then, we retrieved the reported clustering performance of DEC\cite{xie2016unsupervised}, VaDE\cite{jiang2016variational}, ClusterGAN\cite{mukherjee2019clustergan}, and N2D\cite{mcconville2019n2d} from literature as additional baseline results. We report the average clustering performance across four runs of each model, as well as their best clustering performance on each of the benchmark dataset.

We present empirical evidence that consistently shows our method significantly outperforms all our baseline models on each of the benchmark dataset we used. Specifically, our method configurations SNNL-5 and SNNL-7.
\begin{table*}[htb]
    \caption{Clustering Performance on the MNIST Dataset.}
    \label{tab:mnist_clustering_performance_latent_code}
    \centering
    \begin{tabular}{| *{13}{c|}}
        \hline
         \multirow{2}{*}{Method} & \multicolumn{2}{c|}{ACC} & \multicolumn{2}{c|}{NMI} & \multicolumn{2}{c|}{ARI} & \multicolumn{2}{c|}{SIL} & \multicolumn{2}{c|}{CHS} & \multicolumn{2}{c|}{DBI}\\
         \cline{2-13}
         & Average & Best & Average & Best & Average & Best & Average & Best & Average & Best & Average & Best \\ 
         \hline
         Original & 0.525 & 0.547 & 0.497 & 0.499 & 0.367 & 0.367 & 0.078 & 0.078 & 469.231 & 471.494 & 2.605 & 2.592\\
         \hline
         SNNL-2 & 0.549 & 0.552 & 0.512 & 0.52 & 0.387 & 0.393 & 0.086 & 0.089 & 443.535 & 459.947 & 2.631 & 2.619\\
         \hline
         SNNL-4 & 0.562 & 0.569 & 0.516 & 0.521 & 0.4 & 0.409 & 0.086 & 0.089 & 442.462 & 458.644 & 2.640 & 2.583\\
         \hline
         Baseline AE & 0.56 & 0.576 & 0.518 & 0.528 & 0.399 & 0.419 & 0.086 & 0.088 & 445.146 & 457.095 & 2.638 & 2.613\\
         \hline
         SNNL-6 & 0.562 & 0.58 & 0.512 & 0.53 & 0.4 & 0.429 & 0.084 & 0.088 & 437.957 & 449.735 & 2.647 & 2.616\\
         \hline
         SNNL-8 & 0.575 & 0.589 & 0.528 & 0.543 & 0.413 & 0.433 & 0.085 & 0.088 & 437.458 & 447.399 & 2.657 & 2.636\\
         \hline
         DEC\cite{xie2016unsupervised}* & -- & 0.843 & -- & -- & -- & -- & -- & -- & -- & -- & -- & --\\
         \hline
         VaDE\cite{jiang2016variational}* & -- & 0.945 & -- & -- & -- & -- & -- & -- & -- & -- & -- & --\\
         \hline
         N2D\cite{mcconville2019n2d}* & -- & 0.948 & -- & 0.882 & -- & -- & -- & -- & -- & -- & -- & --\\
         \hline
         ClusterGAN\cite{mukherjee2019clustergan}* & -- & 0.95 & -- & 0.89 & -- & 0.89 & -- & -- & -- & -- & -- & --\\
         \hline
         SNNL-1 & 0.901 & 0.952 & 0.862 & 0.881 & 0.849 & 0.898 & \textbf{0.957} & \textbf{0.96} & \textbf{162605.531} & \textbf{214754.189} & 0.245 & \textbf{0.092}\\
         \hline
         SNNL-3 & 0.904 & 0.957 & 0.860 & 0.896 & 0.821 & 0.908 & 0.7 & 0.775 & 14874.154 & 24043.605 & 0.591 & 0.484\\
         \hline
         SNNL-7 & 0.953 & \textbf{0.962} & \textbf{0.891} & \textbf{0.903} & 0.897 & \textbf{0.918} & 0.742 & 0.893 & 24713.490 & 44316.457 & 0.473 & 0.204\\
         \hline
         SNNL-5 & \textbf{0.955} & 0.958 & 0.890 & 0.895 & \textbf{0.903} & 0.911 & 0.874 & 0.887 & 33782.991 & 39916.318 & \textbf{0.239} & 0.186\\
         \hline
    \end{tabular}
\end{table*}

\begin{table*}[htb]
    \caption{Clustering Performance on the Fashion-MNIST Dataset.}
    \label{tab:fmnist_clustering_performance_latent_code}
    \centering
    \begin{tabular}{| *{13}{c|}}
        \hline
         \multirow{2}{*}{Method} & \multicolumn{2}{c|}{ACC} & \multicolumn{2}{c|}{NMI} & \multicolumn{2}{c|}{ARI} & \multicolumn{2}{c|}{SIL} & \multicolumn{2}{c|}{CHS} & \multicolumn{2}{c|}{DBI}\\
         \cline{2-13}
         & Average & Best & Average & Best & Average & Best & Average & Best & Average & Best & Average & Best \\ 
         \hline
         SNNL-8 & 0.52 & 0.548 & 0.561 & 0.569 & 0.385 & 0.395 & 0.120 & 0.125 & 793.908 & 824.502 & 2.143 & 2.103\\
        \hline
         SNNL-2 & 0.535 & 0.553 & 0.568 & 0.576 & 0.397 & 0.406 & 0.118 & 0.123 & 798.417 & 833.797 & 2.141 & 2.088\\
         \hline
         Baseline AE & 0.540 & 0.557 & 0.567 & 0.581 & 0.391 & 0.413 & 0.119 & 0.123 & 806.924 & 830.819 & 2.119 & 2.098\\
         \hline
         Original & 0.518 & 0.559 & 0.518 & 0.531 & 0.366 & 0.389 & 0.186 & 0.191 & 1662.815 & 1673.995 & 1.61 & 1.549\\
         \hline
         SNNL-6 & 0.553 & 0.591 & 0.569 & 0.579 & 0.401 & 0.434 & 0.116 & 0.12 & 796.24 & 808.859 & 2.181 & 2.119\\
         \hline
         SNNL-4 & 0.555 & 0.595 & 0.574 & 0.583 & 0.408 & 0.436 & 0.119 & 0.122 & 790.444 & 803.54 & 2.167 & 2.11\\
         \hline
         ClusterGAN\cite{mukherjee2019clustergan}* & -- & 0.63 & -- & 0.64 & -- & 0.50 & -- & -- & -- & -- & -- & --\\
         \hline
         N2D\cite{mcconville2019n2d}* & -- & 0.672 & -- & 0.684 & -- & -- & -- & -- & -- & -- & -- & --\\
         \hline
         SNNL-3 & 0.672 & 0.693 & 0.682 & 0.691 & 0.527 & 0.535 & 0.575 & 0.59 & 10031.276 & 11013.237 & 0.666 & 0.626\\
         \hline
         SNNL-1 & 0.741 & 0.748 & 0.756 & 0.763 & 0.619 & 0.633 & \textbf{0.963} & \textbf{0.967} & \textbf{212599.963} & \textbf{272354.021} & \textbf{0.331} & \textbf{0.225}\\
         \hline
         SNNL-7 & 0.832 & 0.848 & 0.753 & 0.765 & 0.696 & 0.711 & 0.665 & 0.722 & 10638.715 & 16001.514 & 0.692 & 0.599\\
         \hline
        SNNL-5 & \textbf{0.844} & \textbf{0.856} & \textbf{0.762} & \textbf{0.767} & \textbf{0.714} & \textbf{0.729} & 0.672 & 0.705 & 9646.826 & 10823.988 & 0.628 & 0.546\\
         \hline
    \end{tabular}
\end{table*}

\subsubsection{MNIST}
We had the highest average clustering accuracy of \textbf{95.5\%} on the MNIST dataset using SNNL-7 configuration, and the highest best clustering accuracy of \textbf{96.2\%} using SNNL-5 configuration. Meanwhile, for each of our related models, only their best clustering accuracy was reported. Their results are as follows: DEC had 84.3\%, VaDE had 94.5\%, N2D had 94.8\%, and ClusterGAN had 95\%. If we follow the related works to report the best performance, we outperformed our closest baseline, ClusterGAN, with \textbf{1.2\%} in clustering accuracy. For the full results of clustering performance on the MNIST dataset, we refer the reader to Table \ref{tab:mnist_clustering_performance_latent_code}.
\subsubsection{Fashion-MNIST}
We had the highest clustering accuracy of \textbf{84.4}\% (average) and \textbf{85.6}\% (best) on the Fashion-MNIST dataset using SNNL-5. Meanwhile, our related models had the following results: ClusterGAN had 63\% and N2D had 67.2\%. However, we should note here that N2D was trained on the entire Fashion-MNIST dataset, i.e. both training and test sets. For the full results of clustering performance on the Fashion-MNIST dataset, we refer the reader to Table \ref{tab:fmnist_clustering_performance_latent_code}.
\subsubsection{EMNIST Balanced}
We had the highest clustering accuracy of \textbf{78.5\%} (average) and \textbf{79.2\%} (best) on the EMNIST Balanced dataset using SNNL-5 configuration, while our baseline autoencoder had 35.6\% (average) and 36.1\% (best). Neither the deep clustering algorithms nor N2D used the EMNIST Balanced dataset (or any other EMNIST subsets). For the full results of clustering performance on the EMNIST Balanced dataset, we refer the reader to Table \ref{tab:emnist_clustering_performance_latent_code}.
\begin{table*}[htb]
    \caption{Clustering Performance on the EMNIST-Balanced Dataset.}
    \label{tab:emnist_clustering_performance_latent_code}
    \centering
    \begin{tabular}{| *{13}{c|}}
        \hline
         \multirow{2}{*}{Method} & \multicolumn{2}{c|}{ACC} & \multicolumn{2}{c|}{NMI} & \multicolumn{2}{c|}{ARI} & \multicolumn{2}{c|}{SIL} & \multicolumn{2}{c|}{CHS} & \multicolumn{2}{c|}{DBI}\\
        \cline{2-13}
         & Average & Best & Average & Best & Average & Best & Average & Best & Average & Best & Average & Best \\ 
         \hline
         Original & 0.321 & 0.33 & 0.418 & 0.42 & 0.173 & 0.176 & 0.051 & 0.052 & 237.078 & 237.413 & 2.659 & 2.636\\
         \hline
         SNNL-1 & 0.319 & 0.349 & 0.662 & 0.674 & 0.335 & 0.356 & \textbf{0.897} & \textbf{0.902} & \textbf{71353.661} & \textbf{78749.196} & 1.079 & 1.069\\
         \hline
         SNNL-6 & 0.348 & 0.352 & 0.441 & 0.444 & 0.193 & 0.196 & 0.041 & 0.041 & 222.108 & 225.193 & 2.712 & 2.695\\
         \hline
         SNNL-2 & 0.343 & 0.353 & 0.439 & 0.442 & 0.191 & 0.197 & 0.037 & 0.04 & 223.75 & 224.457 & 2.698 & 2.658\\
         \hline
         SNNL-4 & 0.344 & 0.353 & 0.438 & 0.445 & 0.190 & 0.196 & 0.038 & 0.041 & 225.839 & 227.568 & 2.674 & 2.641\\
         \hline
         SNNL-8 & 0.35 & 0.356 & 0.442 & 0.444 & 0.195 & 0.197 & 0.04 & 0.042 & 223.449 & 225.474 & 2.692 & 2.669\\
        \hline
         Baseline AE & 0.356 & 0.361 & 0.446 & 0.449 & 0.198 & 0.201 & 0.042 & 0.044 & 224.537 & 228.909 & 2.709 & 2.687\\
         \hline
         SNNL-3 & 0.396 & 0.44 & 0.667 & 0.699 & 0.391 & 0.438 & 0.6 & 0.737 & 11138.224 & 16324.421 & 1.378 & 1.345\\
         \hline
         SNNL-7 & 0.701 & 0.743 & 0.753 & 0.775 & 0.529 & 0.634 & 0.580 & 0.761 & 4832.462 & 7098.482 & 0.852 & 0.647\\
         \hline
         SNNL-5 & \textbf{0.785} & \textbf{0.792} & \textbf{0.776} & \textbf{0.783} & \textbf{0.641} & \textbf{0.655} & 0.677 & 0.687 & 4697.866 & 5025.238 & \textbf{0.646} & \textbf{0.607}\\
         \hline
    \end{tabular}
\end{table*}

The full results in Table \ref{tab:mnist_clustering_performance_latent_code}, \ref{tab:fmnist_clustering_performance_latent_code}, and \ref{tab:emnist_clustering_performance_latent_code} not only show that our experimental methods outperformed our baseline methods, but also show that the supervised configurations (SNNL-\{3, 5, 7\}) of the soft nearest neighbor loss performed better than the unsupervised ones (SNNL-\{2, 4, 6, 8\}), with the exception of SNNL-1 on the EMNIST Balanced dataset. This emphasizes the requisite of labels for the soft nearest neighbor loss to illuminate the neighborhood structure of a dataset, thus learning a more clustering-friendly feature representation.

Furthermore, SNNL-\{3, 5, 7\} also outperformed our baseline methods in terms of NMI and ARI. Intuitively, having high NMI and ARI scores for clustering implies that there is a high number of similar data points in each cluster. In other words, similar data points have been correctly assigned to their clusters, which in our case was based on the pseudo-cluster labels we used. We can also see that SNNL-\{1, 3, 5, 7\} had the best scores in terms of SIL, CHS, and DBI. Having a high SIL score implies that the points in a cluster are more similar among themselves than they are to the points from a different cluster. Then, both CHS and DBI measure cluster separation, i.e. CHS defines better separated clusters through variance ratios, while DBI defines better separated clusters through distances among clusters.

Our results support that our method and its variants were able to learn a more clustering-friendly feature representation by having better defined clusters and by having correctly clustered data points, which we visually inspect in the next subsection.

\subsection{Visualizing Disentangled Latent Representation}
\begin{figure}[htb!]
    \centering
    \includegraphics[width=\linewidth]{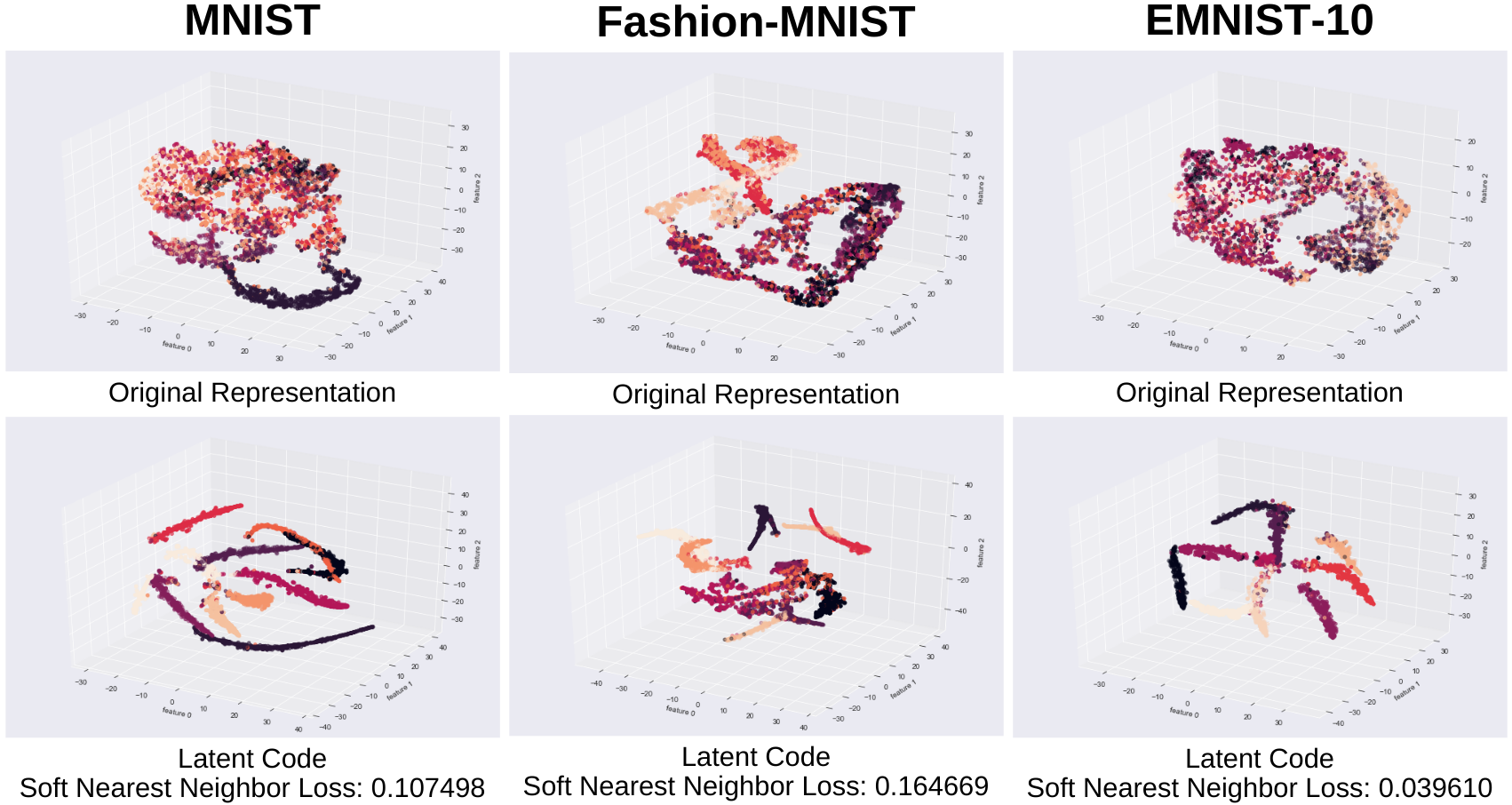}
    \caption{Three-dimensional visualization comparing the original representation and the disentangled latent representation of the three datasets. To achieve this visualization, the representations were encoded using t-SNE with perplexity = 50 and learning rate = 10, optimized for 5,000 iterations, with the same random seed set for all computations. However, for clustering, we used higher dimensionality to achieve better clustering performance. This figure is best viewed in color.}
    \label{fig:embeddings}
\end{figure}
We show the resulting disentangled latent representation for the test set of MNIST, Fashion-MNIST, and EMNIST Balanced datasets in Figure \ref{fig:embeddings}. These latent code representations were obtained after training an autoencoder with SNNL-7 for 50 epochs on a subset of 10,000 training examples each for MNIST and Fashion-MNIST datasets, and on a subset of 20,000 training examples of EMNIST Balanced dataset. However, since the EMNIST Balanced dataset has 47 clusters, we only visualized a randomly chosen 10 clusters for easier and cleaner visualization. We can see in the aforementioned figure that the latent code representation for each dataset indeed became more clustering-friendly by having well-defined clusters (indicated by the cluster dispersion) and correct cluster assignments (indicated by the cluster colors).
\subsection{Clustering on Fewer Labelled Examples}
\begin{figure}[htb]
    \centering
    \includegraphics[width=\linewidth]{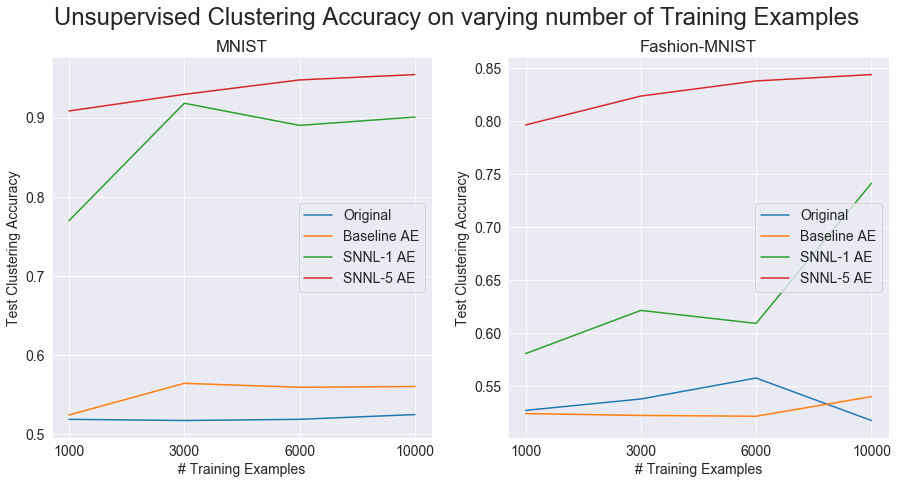}
    \caption{Test clustering accuracy on the MNIST and Fashion-MNIST test sets when small subsets of labelled data are used for training. Both the original representation and the baseline autoencoder do not take advantage of the labelled dataset.}
    \label{fig:clustering_acc_varying}
\end{figure}

\begin{figure}[htb]
    \centering
    \includegraphics[width=\linewidth]{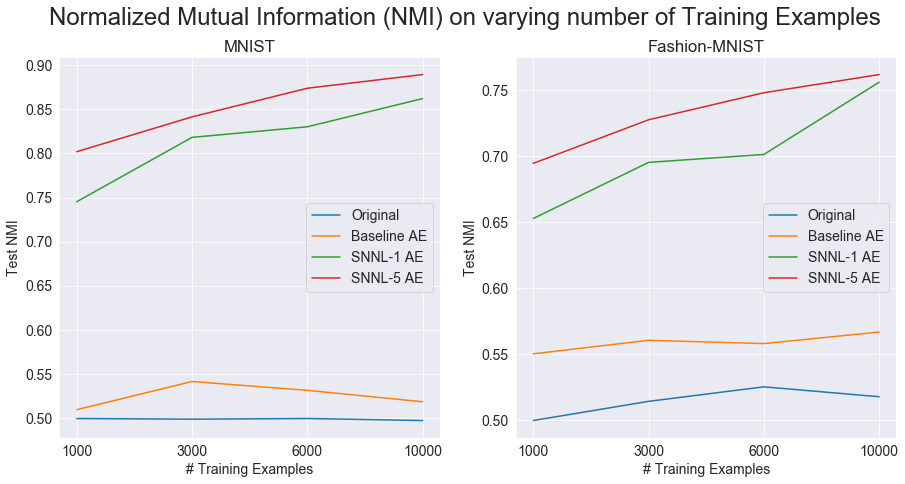}
    \caption{Normalized mutual information (NMI) on the MNIST and Fashion-MNIST test sets when small subsets of labelled data are used for training. Both the original representation and the baseline autoencoder do not take advantage of the labelled dataset.}
    \label{fig:nmi_varying}
\end{figure}
In Figure \ref{fig:clustering_acc_varying}, we show that even with fewer labelled training examples, the clustering accuracy on disentangled latent code representation is still better than on the original feature representation and on the latent code representation from a baseline autoencoder. We trained an autoencoder with SNNL-1 and SNNL-5 configurations on randomly picked 1,000 examples, 3,000 examples, and 6,000 examples from the MNIST and Fashion-MNIST datasets, and treated them as labelled data. Then we evaluated on the full test sets. In contrast, we used the full MNIST and Fashion-MNIST sets for both the original feature representation and the latent code representation from our baseline autoencoder.

On MNIST, our best model configuration was SNNL-5 that had a clustering accuracy of 90.85\%, 92.95\%, and 94.78\% when trained on 1,000 examples, 3,000 examples, and 6,000 examples respectively. Using the same model configuration on Fashion-MNIST, we had a clustering accuracy of 79.63\%, 82.35\%, and 83.78\% when trained on 1,000 examples, 3,000 examples, and 6,000 examples respectively.

The results in Figure \ref{fig:clustering_acc_varying} are average clustering accuracy across four runs of each model on the small labelled subsets of MNIST and Fashion-MNIST datasets. To support these results, we can observe the same trend of clustering performance in terms of NMI in Figure \ref{fig:nmi_varying}. With this, we draw a parallel with the non-linear NCA\cite{salakhutdinov2007learning} performance, where they employed unsupervised pre-training for kNN classification, and found an even better test error rate when they fine-tuned on a small fraction of labelled MNIST dataset -- a test error rate of 1\%.



We argue that this robustness of clustering performance, despite smaller labelled subsets were used for SNNL-trained autoencoders, is due to the soft nearest neighbor loss enabling the learning of a latent code representation that takes into account the distances among pairs of points from the same class or structure relative to points from different classes or structures. In other words, it brings out the neighborhood structure of a dataset.

Finally, we were able to further improve the clustering performance on disentangled latent code representation with our annealing temperature factor. The intuition as to how this annealing factor improves the soft nearest neighbor loss builds on the results in Figure \ref{fig:synthetic_snnl}, and may be described as follows: as the training progresses, the data points of similar class or structure become closer. Hence, there are fewer widely separated points of the same class or structure as training progresses. Consequently, this renders the use of high temperatures less relevant over time, since the class- or structure-similar points are already becoming entangled -- which in turn, makes the latent code representation disentangled.

\section{Conclusion}
Compared to deep clustering methods\cite{jiang2016variational, mukherjee2019clustergan, xie2016unsupervised}, we employed a simpler approach to cluster the latent code representation from an autoencoder. We used a composite loss of the autoencoder reconstruction loss and the soft nearest neighbor loss to learn a more clustering-friendly latent code representation from an autoencoder, thereby improving the k-Means clustering performance on our datasets. We expanded the soft nearest neighbor loss by introducing an annealing temperature factor, which led to an even better disentanglement and k-Means clustering performance. We posit that our annealing mechanism helps by adapting the temperature needed throughout a training.
Using our approach, we had a clustering accuracy of \textbf{95.5\%} (96.2\% on best run), \textbf{84.4\%} (85.6\% on best run), and \textbf{78.5\%} (79.2\% on best run) on the MNIST, Fashion-MNIST, and EMNIST Balanced datasets respectively, outperforming all our baseline models.

\end{document}